\documentclass[11pt,a4paper]{article}
\usepackage[hyperref]{emnlp2020}
\usepackage{times}
\usepackage{latexsym}
\usepackage{microtype}
\usepackage{amsmath}
\usepackage{amssymb}
\usepackage{latexsym}
\usepackage{mathtools}
\usepackage{algorithm,algpseudocode}
\usepackage{bbm}
\usepackage{url}
\usepackage{xcolor}
\usepackage{tabularx}
\usepackage{multirow}
\usepackage{booktabs}
\usepackage{xspace}
\usepackage{kotex}
\usepackage[page]{appendix}
\usepackage{enumitem}

\usepackage{subfig}

\usepackage{microtype}

\aclfinalcopy 
\def\aclpaperid{2612} 

\newcommand{\method}{\textsc{Inspired}\xspace}
\usepackage{etoolbox}
\usepackage{pifont}
\usepackage{xcolor}
\newcommand{\cmark}{\textcolor{green!80!black}{\ding{51}}}
\newcommand{\xmark}{\textcolor{red}{\ding{55}}}
\usepackage{array,booktabs}
\usepackage{fdsymbol}
\usepackage{graphicx}
\newcommand{\heart}{\ensuremath\heartsuit}

\usepackage{cleveref}
\usepackage{commath}
\usepackage{fdsymbol}
\newcommand\tab[1][5mm]{\hspace*{#1}}
\crefformat{section}{\S#2#1#3} 
\crefformat{subsection}{\S#2#1#3}
\crefformat{subsubsection}{\S#2#1#3}

\title{\textsc{Inspired}: Toward Sociable Recommendation Dialog Systems}

\author{Shirley Anugrah Hayati$^\heart$ \tab Dongyeop Kang$^\largewhitestar$\tab  Qingxiaoyang Zhu$^\heart$ \\ \textbf{Weiyan Shi}$^\heart$ \tab \textbf{Zhou Yu}$^\heart$\\
  $^\heart$Department of Computer Science, University of California, Davis\\
  $^\largewhitestar$University of California, Berkeley\\
  \texttt{\{sahayati, qinzhu, wyshi, joyu\}}\texttt{@ucdavis.edu} \\
  \texttt{dongyeopk@berkeley.edu}
 }

\date{}

\begin{document}
\maketitle
\begin{abstract}
In recommendation dialogs, humans commonly disclose their preference and make recommendations in a friendly manner. However, this is a challenge in developing a sociable recommendation dialog system, due to the lack of dialog dataset annotated with such sociable strategies. 
Therefore, we present \textsc{Inspired}, a new dataset of 1,001 human-human dialogs for movie recommendation with measures for successful recommendations. To better understand how humans make recommendations in communication, we design an annotation scheme related to recommendation strategies based on social science theories and annotate these dialogs. Our analysis shows that sociable recommendation strategies, such as sharing personal opinions or communicating with encouragement, more frequently lead to successful recommendations. Based on our dataset, we train end-to-end recommendation dialog systems with and without our strategy labels. In both automatic and human evaluation, our model with strategy incorporation outperforms the baseline model. This work is a first step for building sociable recommendation dialog systems with a basis of social science theories\footnote{Dataset and code are available at \url{https://github.com/sweetpeach/Inspired}}.
\end{abstract}

\section{Introduction}

\begin{figure}[t!]
\vspace{-2mm}
\centering
{
\includegraphics[trim=0cm 15cm 0cm 0cm,clip,width=0.95\linewidth]{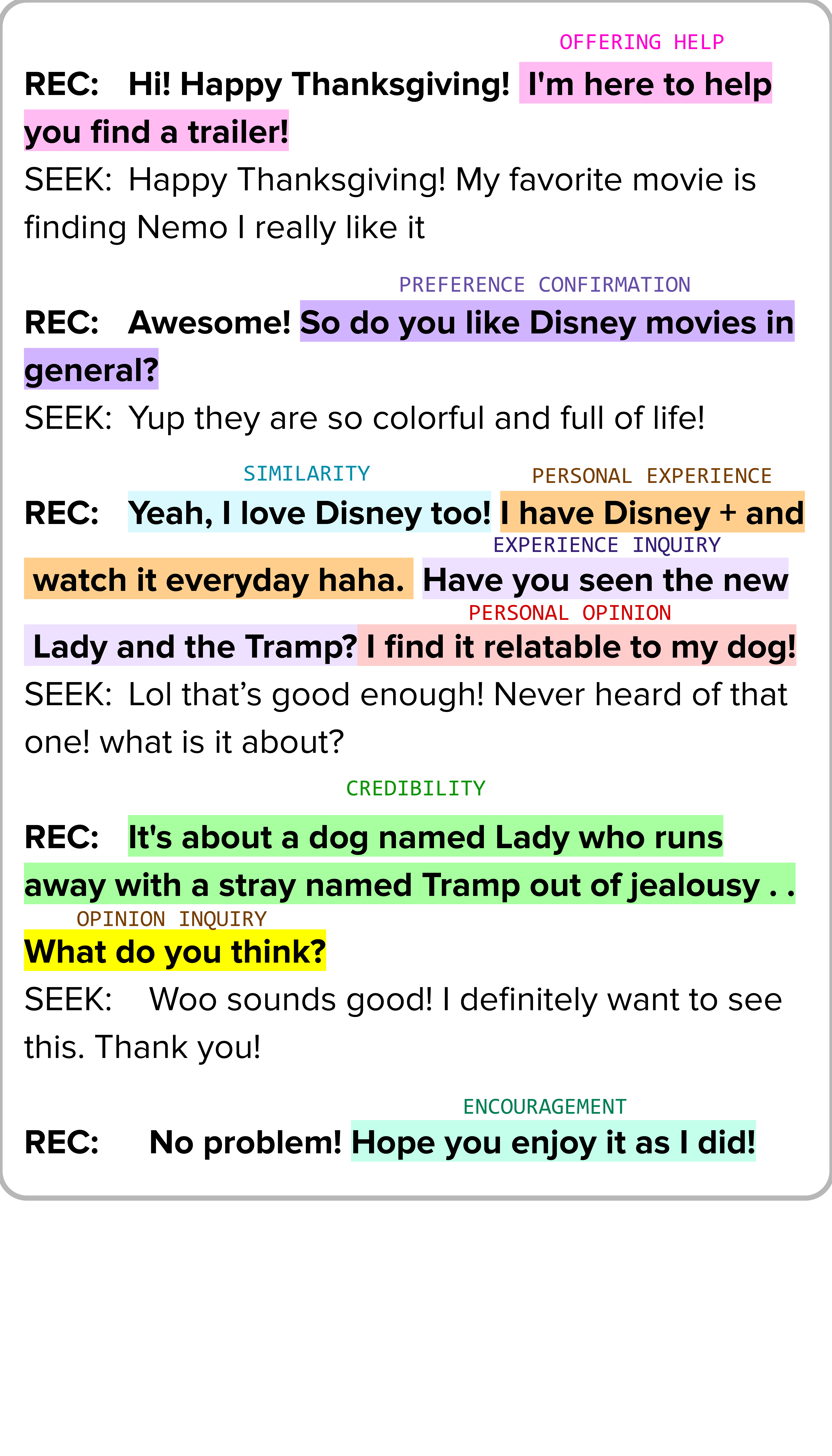}
}
\caption{
\label{fig:example_dialog_with_annotation}  An example snippet of human-human recommendation dialog in \method. 
REC refers a person who recommends a movie and SEEK refers a person who looks for a recommendation. Above each recommender's utterance is the recommendation strategy annotated by human workers. Best seen in colors. 
}
\vspace{-5mm}
\end{figure}

\begin{table*}[ht!]
\centering
\begin{tabular}{@{}p{3cm}cccc@{}}
\toprule
\multirow{2}{*}{\textbf{Dataset}} & \multirow{2}{*}{\textbf{\textsc{Inspired}}} & \textsc{ConvRec} & \textsc{GoRecDial} & \textsc{ReDial}\\
 & & \cite{sun2018conversational} & \cite{kang2019-recommendation}& \cite{li2018towards} \\
\midrule
Naturalness &  \cmark & \xmark & \xmark & \cmark\\ 
Sociable Strategies &  \cmark & \xmark & \xmark & \xmark\\ 
Movie Information  & \cmark & \xmark & \cmark & \xmark\\
Conversation Types  & Mixed & QA & Mixed & Mixed\\
\#Dialogs  & 1,001 & 385 & 9,125 & 10,006\\
\#Utterances  & 35,811 & - & 160,904 & 182,150\\
\bottomrule
\end{tabular}\vspace{-2mm}
\caption{\label{tab:comparison_table} Comparison of related recommendation dialog datasets. ``QA'' refers to question-answer pairs. ``Mixed'' indicates that the conversations contain both statements and question-answer pairs. \textsc{ConvRec} collected 385 human-curated dialogs, but only released 875,721 simulated dialogs. }
\vspace{-2mm}
\end{table*}
Sociable conversational agents build rapport with
users, in order to gain trust and favor from them. Social science researchers believe that the rapport influence a more persuasive recommendation to successfully suggest an item that satisfies user needs \cite{Yoo2012,gkikapersuasive,Pecune2019, gretzel2016}. 

However, existing works on recommendation dialog systems lack a study about communication strategies used by human speakers for making successful and persuasive recommendations.
They collect the dataset in scenario-based settings or convert product review datasets into question-answering conversations \cite{reschke2013, yan2017building,sun2018conversational, kang2019-recommendation, li2018towards}.
Common issues with these types of datasets are: (1) homologous utterances, (2) mostly question-answering pairs, and (3) lack of user engagement. 

In this work, we aim to validate whether sociable recommendation strategies are effective for making a successful recommendation in a dialog. To do so, we propose \textsc{Inspired}, a recommendation dialog dataset of two-paired crowd-workers in a natural setting, with additional annotations for sociable recommendation strategies. 
The dataset consists of 1,001 dialogs, and each utterance is manually annotated with the sociable strategies based on social science theory. 
To encourage more natural dialog flow, we do not set any restrictions on the number of movies or the type of movies to recommend. 
Figure \ref{fig:example_dialog_with_annotation} shows an example of annotated dialog. More examples are in Table \ref{tab:example_human_human_dialogs} and \ref{tab:example2_human_human_dialogs} in the Appendix.

Our analyses show that sociable recommendation strategies are correlated with successful recommendation in dialogs. These insights motivate us to build a more sociable recommendation dialog system to achieve better persuasion outcomes. 

For extrinsic evaluation, we build two end-to-end dialog systems trained on the \textsc{Inspired} dataset: one is encoded with recommendation strategies and the other is not. We find that the model encoded with our strategy annotations performs better in both automatic and human evaluation.

We believe that enriching the intersection between social science and computational linguistics in \textsc{Inspired} opens plenty of rooms for future studies on sociable recommendation dialog.

\section{Related Work}

\paragraph{Social science theories on recommendation.}
Psychological researchers believe that interactions with recommendation systems should not only be seen  from a technical perspective but should also be examined  from a social and emotional perspective \cite{zanker2006persuasive}.
\citet{Yoo2012} propose that credibility, likeability, friendliness, humor, and other language styles are significant factors for persuasive recommendations. 
\citet{Pecune2019} has studied modeling social explanation for movie recommendation, such as personal opinion and personal experience. 
\citet{haubl2003preference} find that more information on recommendation may help consumers make better purchase decisions, but leave them overwhelmed with the abundant information. 
Inspired by these theories, we borrow such principles in the design of our sociable recommendation strategies. 

\paragraph{Conversational recommendation systems.} 
While studies on conversational recommendation systems have been done, none of them focus on the sociable recommendation strategies for persuasive outcome. This is is due to the lack of existing datasets for studying effective strategies in recommendation dialog. Table \ref{tab:comparison_table} compares different factors across the recommendation dialog datasets including \method.

Prior works on recommendation dialogs collect data based on template-based question-answering pairs from user reviews \cite{thompson2004personalized, reschke2013, sun2018conversational, zhang2018towards}. These datasets contain structured utterances where the recommender continuously asks for the seeker's product preference. 

\citet{kang2019-recommendation} collected goal-driven recommendation dialogs (\textsc{GoRecDial}) in a gamified setting where both speakers are given a small set of movies with descriptions to find the best recommendation. This role-play game setting may not effectively reflect the real-world situation since the seeker pretends that they like the given movies. 

The most similar work to ours is \citet{li2018towards}'s \textsc{ReDial} dataset which consists of chit-chats for movie recommendation. However, the recommendations are conditioned on the movies mentioned in the dialog, and not directly on the language usage. Also, they tend to mention only movie names rather than an in-depth discussion on the movie preference. 

Our work is also closely related to \citet{radlinski2019} on movie preference elicitation and \citet{galetzka2020} on movie discussion in the dialog setting. Preference elicitation is an important step for the human recommender to comprehend seeker's taste before recommendation, but these datasets are not recommendation conversations. 

Meanwhile, dialogs in \method have both stages: preference elicitation and recommendation. \method also captures sociable recommendation strategies in conversations and measures recommendation with ratings. 

\begin{figure}[t]
\vspace{0mm}
\centering
{
\includegraphics[trim=0cm 0.3cm 0cm 0cm,clip,width=0.9\linewidth]{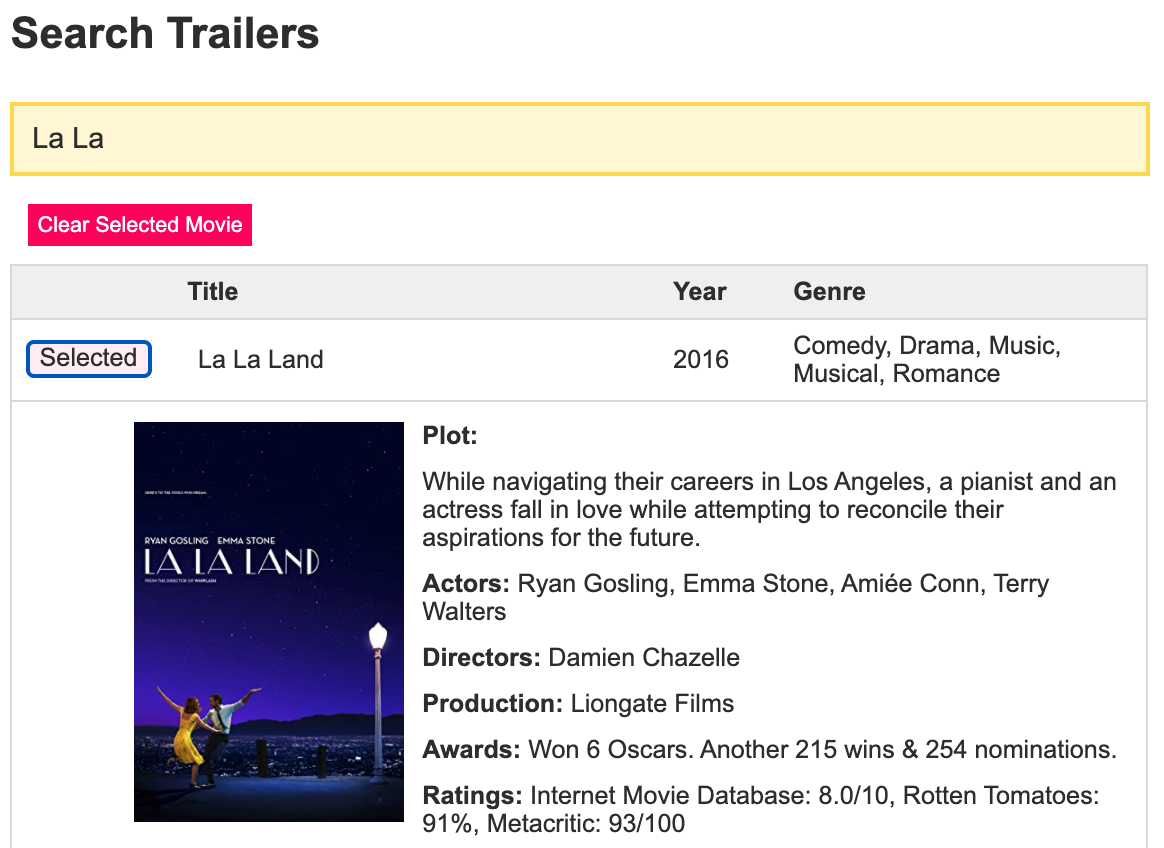}
}
\caption{\label{fig:snapshot_movie_database} Movie search interface for recommenders. 
\vspace{-1mm}
}
\vspace{-2mm}
\end{figure}

\paragraph{Sociability in dialog systems.} 
In human-human conversations, people engage in a talk that does not only contain task-oriented topics \cite{bickmore2005social}. Thus, sociability has raised more attention in dialog systems as they become more sociable, engaging, and user-adaptive  \cite{zhang2018personalizing, shi2018sentiment, goker2000personalized}.

\citet{zhang2018personalizing} proposed a chit-chat dataset  and presented the task of more personalized dialogs system  conditioned on user profile information. 
Sociability leads to a more persuasive conversation \cite{Yoo2012}, so social skills are essential for dialog systems to make successful recommendations. 

Communication strategies on specific tasks, such as donation and product price negotiation, have been found useful for task completion \cite{wang2019-persuasion, zhou-etal-2019-dynamic}. 
In this work, we connect different sociable strategies with recommendation in dialog settings and show that sociable strategies have a positive impact on recommendation success.




\section{Recommendation Dialog Collection}
\subsection{Movie Database Creation}
To ensure that the recommended movie has trailers and metadata information, we curate a database with all movie trailers from Movieclips Trailer\footnote{\url{youtube.com/user/movieclipsTRAILERS}} released between 2008 and 2020, and movies from MovieLens dataset \cite{Harper2015}. In total, we have 17,869 movies with trailers and metadata information. We design a simple movie search interface (Figure \ref{fig:snapshot_movie_database}) to assist recommenders in searching for a movie. 
\subsection{Recommendation Task}
We recruit crowd-workers from Amazon Mechanical Turk. In each conversation, two workers are randomly paired and assigned different roles: one as a recommender and another as a seeker. Our collection set-up is more realistic compared to prior works as (1) recommenders have no limitations of the number of movies to recommend, (2) seekers accept or reject a movie following their true preference, and (3) we record if seekers actually watch the video trailer or not. 

\paragraph{Recommender.} 
Recommenders' task is to recommend a movie successfully to the seeker. Before chatting, we show them tips for sociable recommendation strategies with example utterances. Then they chat with the seekers in two phases: user information gathering and movie recommendation. In the user information gathering phase, recommenders are asked to understand the seekers' movie tastes. In the recommendation phase, the recommenders can still request seekers' preference while browsing movies to recommend. We encourage the recommenders to continue the conversation until seekers accept a movie. 

\paragraph{Seeker.} 
Seekers are asked to talk about movie recommendations without any strategy support. After they complete the conversation, seekers can opt to accept or reject the provided movie recommendations. If the seekers accept the recommendation, they can watch the entire recommended movie trailer or part of it, or simply skip it after the conversation. We record how long seekers watched the recommended movie trailer and ask them to rate the trailer on 5-Likert scale in the post-task survey. 

\begin{table}
\centering
\begin{tabular}{lr}
\toprule
\multicolumn{2}{@{}l}{\textbf{Dataset Statistics} }  \\\hline 
\# Dialogs & 1,001 \\ 
\# Utterances  & 35,811\\
Average turns per dialog  & 10.73\\
Average tokens per utterance & 7.93\\ 
\# Unique tokens & 18,316\\ 
\midrule
\multicolumn{2}{@{}l}{\textbf{Recommender's Statistics}}  \\
\midrule
\# Utterances & \multicolumn{1}{r}{18,339} \\
Average tokens per turn & \multicolumn{1}{r}{14.64} \\
\# Unique tokens & \multicolumn{1}{r}{13,753} \\
\midrule
\multicolumn{2}{@{}l}{\textbf{Seeker's Statistics}} \\
\midrule
\# Utterances & \multicolumn{1}{r}{17,472} \\
Average tokens per turn & \multicolumn{1}{r}{12.12} \\
\# Unique tokens &
\multicolumn{1}{r}{10,097}\\
\bottomrule
\end{tabular}
\caption{\label{tab:data_statistics} \textsc{Inspired}'s statistics. \# denotes the number.}
\label{tab:stats}
\vspace{-2mm}\vspace{-2mm}
\end{table}

\subsection{Dialog Data Collection Details}
We use ParlAI platform \cite{miller2017parlai} and hire 1,594 US crowd-workers from Amazon Mechanical Turk with a minimum of 90\% task acceptance rate. 
The dialog collection process lasted from November 2019 to March 2020. 

Workers first fill out questionnaires related to their personality traits and values before their conversations. The questionnaire consists of three personality trait models: the Big Five personality traits (15 questions) \cite{goldberg1993structure}, the Schwartz Portrait Value (10 questions) \cite{schwartz2003proposal}, and the Decision Making Style (2 questions) \cite{hamilton2016development}\footnote{We also release this personality information in our dataset for future work}. Then, recommenders start the conversation and both workers should chat for a minimum of 10 turns or until a recommendation is made.  After the conversation ends, both workers will answer a post-task survey of demographic questions such as age, and gender. Seekers are asked to rate the trailer with a high score (4 or 5 stars) on a 5-Likert scale and provide the reason of why they reject or do not finish watching the video. Both workers receive a bonus up to \$2 if they complete the entire process in addition to the base pay of \$0.5. 



Table \ref{tab:data_statistics} presents statistics of the collected dataset\footnote{Dialog collection interfaces are in \cref{interfaces} in Appendix}. Even though our dataset has relatively small number of samples compared to \textsc{ReDial} or \textsc{GoRecDial}, it has human annotations on each sociable strategy. Moreover, our dataset can be used in combination with other datasets in a semi-supervised setting, as shown in our implementation of recommendation dialog systems in \cref{recdialsys}.

\begin{table}
\centering
\begin{tabular}{l|rr}
\textbf{Cases}& \multicolumn{2}{c}{\textbf{\#Dialogs}}  %
 \\ \midrule
Accept (Rating 4-5) & 532& (53.1\%)\\
Accept (Rating 3 or lower) & 45 &(4.5\%)\\
Accept (Other Reasons) & 289 & (28.9\%) \\
Accept Uninterested & 123 & (12.3\%)\\
Reject & 12 & (1.2\%)\\
\end{tabular}
\caption{\label{tab:cases} Statistics of dialogs when the seekers accept or reject the final recommended movie. ``Accept (Rating 4-5)'' means that the seekers accept the recommendation and give rating 4 or 5, and the same is for ``Accept (Rating 3 or lower)''. ``Accept (Other Reasons)'' suggests that the seeker gives other reasons for not finishing the video. ``Accept Uninterested'' indicates that the seekers accept the recommendation, do not finish watching the video, and explains in the post-task survey that they are not interested in the recommended video.}
\end{table} 

The statistics of accept and reject cases are shown in Table \ref{tab:cases}. We have higher number of successful cases (79.7\%) compared to failure cases. This shows that people tend to accept recommendations, and it is not surprising since watching a video trailer is an entertaining, low-risk activity. For training the dialog model, we use every dialog from all cases so that the dialog system will be able to respond to diverse responses. 

\begin{table*}[t]
  \centering
  \begin{tabularx}{0.99\textwidth}{@{}r X@{}}
  \toprule
    \textbf{Category} & \textbf{Example}\\
    \midrule
    \textsc{Personal Opinion} &  ``I really like Disney's more recent princesses''
    \\ 
    \textsc{Personal Experience} & “I have Disney+ and watched it everyday!” \\
    \textsc{Similarity} & ``Oh, I love Disney as well.'' \\
    \textsc{Encouragement} & ``You should definitely watch it!''\\ 
    \textsc{Offering Help} &  ``I'm here to help you find a trailer!'' \\ 
    \textsc{Preference Confirmation} & ``So do you like Disney movies in general?''\\ 
    \textsc{Credibility} &  ``It's about a dog named Lady who runs away with a stray named Tramp''\\
    \textsc{Self-Modeling} &  ``We are planning to go see Maleficent, we heard it was a very good movie.''\\
    \textsc{Experience Inquiry} &  ``Have you seen the new Lady and the Tramp?''\\ 
    \textsc{Opinion Inquiry} &  ``What do you like about the Avengers: End-game?''\\ 
    \textsc{Recommendation} &  ``You should check out Shazam!''\\
    \bottomrule
  \end{tabularx}
  \caption{Example utterances for each strategy.}
  \label{tab:annot_example_sentences}
  \vspace{-2mm}
\end{table*}

\section{Recommendation Strategy Annotation}
\subsection{Strategy Definition}
\label{strategy_def_section}
After conversations are collected, two experts, trained with linguistics background, develop an annotation scheme using content analysis method \cite{krippendorff2004reliability} and from past study on human behavior in making recommendations. Similar approaches have been done in prior studies on work for persuasion task \cite{wang2019-persuasion} or negotiation task \cite{ zhou-etal-2019-dynamic}. We divide the recommendation strategies into two categories: sociable strategies and preference elicitation strategies. Sociable strategies are also derived from our literature study on the social science theories. 

\textbf{Sociable strategies} contain eight strategies related to the recommendation task. These strategies relate to the recommenders trying to build rapport with the seekers.
\begin{itemize}[leftmargin=*]
 \item \textbf{Personal opinion} refers to a condition when recommenders express their subjective opinion about a movie, including its plot, actors, or other movie attributes. 
 
 \item \textbf{Personal experience} refers to the use of sharing personal experience related to a movie. For example, recommenders may say that they watch the movie several times to convince the seekers that the movie is good. Both personal opinion and personal experience are part of self-disclosure that leads to establishing rapport with the seekers \cite{altman1973reciprocity}. 
 
 \item \textbf{Similarity} refers to a condition when the recommenders are empathizing and being like-minded toward seekers about their movie preference to produce similarity among them. Similarity is believed to influence the seekers’ liking for the source that leads to trust the recommenders' judgment more \cite{okeefe2004}, following \citet{Lazarsfeld1964FriendshipAS}'s homophily theory that states humans like other people who are similar to them.
 
 \item \textbf{Encouragement} is the use of praise of the seekers' movie taste and encouragement to watch a recommended movie to build rapport and promote the recommended movie.
 
 \item \textbf{Offering help} is a strategy when the recommenders disclose explicit intention to help the seeker or being transparent. It is a part of ``transparency'' strategy from  \citet{gretzel2016}.

\item \textbf{Preference confirmation} is a strategy when the recommenders ask or rephrase the seeker's preference. This strategy is also a part of ``transparency'' strategy which states that the recommenders disclose their thinking process of understanding the seekers' preference. 

\item \textbf{Self-modeling} is a strategy when the recommender becomes a role model to do something first so that the Seeker would follow \cite{dowrick1999}. 
\item \textbf{Credibility} happens when the recommender shows expertise and trustworthiness in providing information to persuade the seeker \cite{Fogg2002, okeefe2004, rhoads2002}. In our study, a recommender is doing credibility appeal when they provide factual information about movie attributes, such as the plot, actors, or awards that the movie has.
\end{itemize}

\textbf{Preference elicitation inquiries} include the following inquiries that are asked by the recommenders to know the seekers' movie tastes. 

\begin{itemize}[leftmargin=*]
 \item \textbf{Experience inquiry} asks for seeker's  experience on movie watching, such as whether a seeker has watched a certain movie or not.
 \item \textbf{Opinion inquiry} asks for seeker’s opinion on movie-related attributes. Example answers for this inquiry is the seeker's explanation on what they like about the plot or if they admire the actors' acting skill.
\end{itemize}

Other kinds of utterances, such as greetings or thanks, fall into non-strategy category. We also label sentences which are recommendation. Recommendation is defined as when the recommender suggests a new movie title for the first time for the seeker. 30\% of the recommendation sentences are ``experience inquiries'', 27\% are ``encouragement'', and 14\% are ``personal opinion''. Example annotated utterances are displayed in Table \ref{tab:annot_example_sentences}. Meanwhile, Table \ref{tab:annot_data} shows the number of annotated utterances in  \textsc{Inspired}.

\begin{table}
\centering
\begin{tabular}{@{}lrr@{}}
\toprule
\textbf{Category} & \multicolumn{2}{c}{\textbf{\#Utterances}}\\ \midrule
\multicolumn{3}{@{}c@{}}{\textbf{Sociable Strategies}}\\ \midrule
Credibility  & 2,687 &(13.7\%)\\ 
Personal Opinion  & 2,599 &(13.9\%)  \\
Encouragement & 1,975 &(10.6\%)\\
Similarity  & 957 &(5.1\%)\\
Offering Help & 953 &(5.1\%)\\
Preference Confirmation &  950& (5.1\%)\\
Personal Experience  &  564 &(3\%)\\
Self-Modeling &  449 &(2.4\%)\\ \midrule
\multicolumn{3}{c}{\textbf{Preference Elicitation Inquiries}} \\ \midrule
Experience Inquiry & 1,505 &(8.1\%)\\
Opinion Inquiry &  2,120 &(11.3\%)\\\midrule
\multicolumn{3}{@{}c@{}}{\textbf{Non-strategy}} \\ \midrule
No Strategy  & 2,566 & (13.7\%)\\
Acknowledgment  & 1,354 &(7.2\%)\\\midrule
Recommendation  & 2177 &(6.1\%)\\\bottomrule
\end{tabular}
\caption{Statistics of the number of utterances annotated with strategies in \textsc{Inspired}.}
\label{tab:annot_data}
\vspace{-4mm}
\end{table}

\subsection{Annotation Quality}
To ensure annotation quality, we separate our annotation study in two steps. First, we hire two experts with linguistics training to perform annotation, in order to test the validity of the scheme. The two experts annotated 30 randomly selected conversations and reached a Kappa agreement of 0.77, suggesting that our scheme is possible to replicate. 

Our dataset contains more than 18k utterances, so it's too costly to hire experts to annotate all of them. In the second step, We hire US-based crowd-workers (95\% task acceptance) from Amazon Mechanical Turk for the annotation tasks. In each task, a worker was given a tutorial of the annotation and then they were given 10 dialogs to annotate. One of the dialogs was labeled by experts to calibrate the quality of the worker's annotation, called as evaluation dialog. Five workers work on the same task. We filter out workers whose score is below the threshold 0.60 on the evaluation dialog. To set this threshold in a reasonable value, we conducted the following study. This time we ran onetask in which all the dialogs are already labeled with the experts including the evaluation dialogs. We found that if the workers’ score on the evaluation dialog is above 0.60, their agreement score with the expert’s annotation on the rest of the dialogs in this task is 0.77.

These selected high quality crowd-workers annotate the rest of the dialogs. We still have five workers annotate the same dialog. If more than one worker disagrees on a utterance's annotation, the experts are then involved to annotate them as quality control. The inter-annotator majority agreement among all workers is 0.78 over all dialogs. The annotation scheme for the crowd-workers are provided in Figure \ref{fig:strategy_definition} in the Appendix. 

\begin{figure}
\vspace{0mm}
\centering
{
\includegraphics[width=\linewidth]{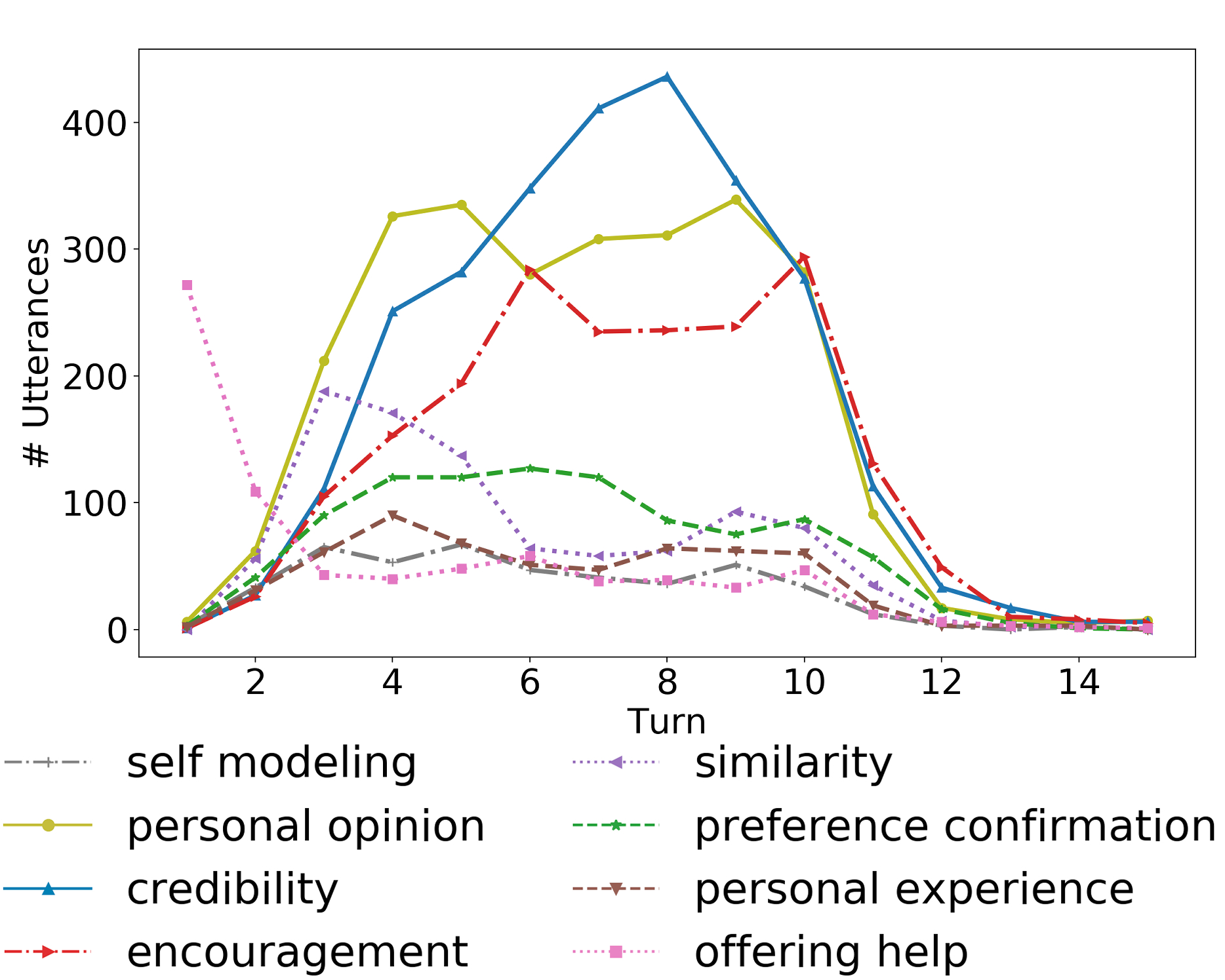}}
\caption{\label{fig:strategy_turn} Distribution of sociable strategies over the dialog turns. 
Best viewed in color.}
\end{figure}

\section{Strategy Analyses}
\subsection{Distribution of Strategies over Dialog}
As shown in Figure \ref{fig:strategy_turn}, we observe that different sociable strategies are unequally distributed across conversation turns. 
Most notably, ``offering help'' and ``similarity'' often happen at the beginning, indicating that recommenders strategically attempt to build rapport with seekers at the early stages. Then, ``credibility'' and ``personal opinion'' frequently appear in the conversations, as recommenders seek to persuade. Moreover, ``encouragement'' mostly appears in the middle and at the end of conversations. 

\subsection{What Strategies Contribute to Successful Recommendations?}

We study the association of sociable strategies and successful recommendations. 
A recommendation is considered successful if seekers finish watching a substantial portion of the recommended movie trailer and rate the trailer with a high score (4 or 5 stars). We set a threshold that seekers need to watch at least more than 50\% of the video duration since some videos have advertisements at the end, etc. On the other hand, a recommendation is considered unsuccessful if the seekers reject the recommendation (``Reject'') or skip watching the trailer (``Accept Uninterested''). Thus, for our analysis, we use 532 successful dialogs and 135 unsuccessful dialogs for our analysis on association of strategies in successful recommendations. 

To analyze the effect of our sociable recommendation strategies on success of recommendation, we run a logistic regression model to predict the success of recommendation (1 = successful, 0 = unsuccessful). We use frequency of the strategy in a dialog as the feature value.

Table \ref{tab:strategy_success} shows the coefficients of each strategy with respect to the recommendation.
We observe that ``personal opinion'', ``similarity'', ``encouragement'', and ``credibility'' strategies have a significant positive effect on successful recommendations. This confirms with the previous studies that more sociable recommenders are more likely to be successful in the recommendation.

``Similarity'' strategy has the highest coefficient value which suggests that if the recommender is conforming to the seeker's preference, the seeker is more likely to favor the recommendation. This also supports the theory in \citet{okeefe2004} that likeability helps in recommendation. We also observe that all the preference elicitation inquiries are not significantly contributing to the successful recommendation. From this result, we are not saying that recommenders need not to query seekers' preferences since it is crucial to understand their tastes. However, a more sociable approach is necessary for a more successful recommendation. 

\begin{table}
\centering
\begin{tabular}{@{}lr@{}}
\toprule
\textbf{Category} & \textbf{Coefficient}\\ \midrule
\multicolumn{2}{@{}c@{}}{\textbf{Sociable Strategies}}\\ \midrule
Personal Opinion  & 0.12* \\
Personal Experience  & 0.05  \\
Similarity  & \textbf{0.23}* \\
Encouragement & 0.20** \\
Offering Help & 0.03 \\
Preference Confirmation & 0.05\\
Self-Modeling & 0.02\\ 
Credibility  & 0.09* \\\midrule
\multicolumn{2}{@{}c@{}}{\textbf{Preference Elicitation}} \\ \midrule
Experience Inquiry & $-$0.01  \\
Opinion Inquiry & 0.06 \\
\bottomrule
\end{tabular}
\caption{Associations between different strategies and successful recommendation.  *$p <$ 0.05, **$p <$ 0.01}
\label{tab:strategy_success}
\end{table}

\subsection{Are Sociable Strategies Still Significant with the Presence of Movie Attributes?}
In a recommendation task, a natural question to ask is how big a role the recommended product plays in the acceptance of recommendation. If the quality of the product matters  more than how you recommend, it makes more sense to improve the products rather than the recommendation skills.
Therefore, we also analyze if adding movie attributes, such as the genre, recent release date, and the number of likes of the movie trailer have an impact on successful recommendation along with the eight sociable strategies and two preference elicitation inquiries. 

For the popularity, we categorize the top 10\%  movies in terms of the number of  likes  to be popular and the rest to be  non-popular in our database. A movie is said to be recent if it is released in 2019 or 2020. For the genre, we select the top five most popular genres in the movie database. When we check with the recommended movies in \method,  96\% of recommended movies are covered by the top five genres. 

Results of the analysis between the strategies and movie attributes are shown in Table \ref{tab:movie_attribute} in the Appendix. Sociable strategies remain significantly correlated with successful recommendations. Recommenders who perform ``similarity'' strategy, express ``personal opinion'', and show ``encouragement'' are more likely to successfully recommend a movie ($p < 0.05$). Surprisingly, none of the movie attributes has significant effect on successful recommendations. A possible reason is that the seekers' movie tastes are so diverse that movie attributes such as genre do not have a significant impact on the recommendation success. 

\section{Recommendation Dialog Systems}
\label{recdialsys}
To evaluate how the strategies in \textsc{Inspired} are useful in creating a more engaging and persuasive recommendation dialog, we develop a generative dialog model as our baseline to compare against our strategy-incorporated dialog system. We split the dialogs into 801/100/100 for train/validation/test split. We use external recommendation system from TMDB\footnote{\url{https://www.themoviedb.org/}} with heuristics to select the movies. More details for heuristics and training set-up are in the Appendix. 

\subsection{Baseline  Model}
\begin{figure}
\centering
{
\includegraphics[trim=0cm 0.3cm 0cm 0cm,clip,width=\linewidth]{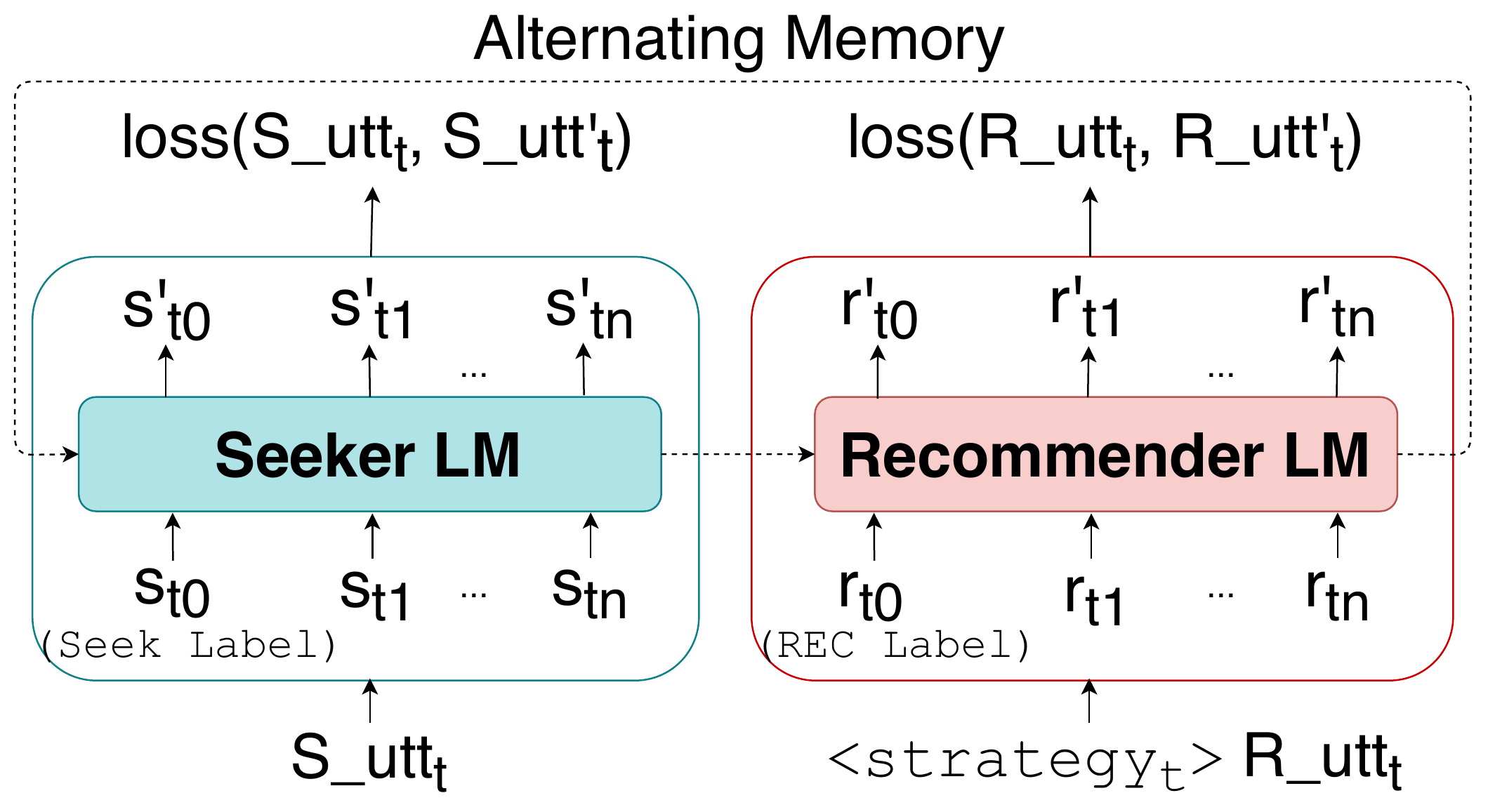}
}
\caption{The Seeker's language model (Seeker LM) and the Recommender's language model (Recommender LM) are separate memory. The Seeker LM input at turn $t$ is the seeker's utterance $S\_utt_t$ consisting of a sequence of tokens $s_{t0}, s_{t1},... s_{tn}$. The Recommender LM input at turn $t$ is the recommender's utterance $R\_utt_t$ consisting of a sequence of tokens $r_{t0}, r_{t1},..., r_{tn}$. The  $<$\texttt{strategy$_t$}$>$ prepended as a special token. For the baseline, the recommender's input does not contain the strategies. }
\label{fig:diagram}
\end{figure}

The baseline dialog model uses two separate Transformer-based pretrained language models \cite{vaswani2017, radford2019language, ardm2019} to learn the recommender's and seeker's language models separately in alternating order. 
Both language models are trained to maximize the likelihood of generating ground truth utterance on the alternating memory as shown in Figure \ref{fig:diagram}. The model is pretrained on non-task related corpus, WebText, and task-related corpus: recommendation dataset from \textsc{ReDial} \cite{li2018towards} and movie preference elicitation dataset ~\cite{radlinski2019}. Then, we fine-tune the model with \textsc{Inspired}. 

We replace movie attributes such as titles, actors, and genres with indexed placeholders. It is because in a single conversation, multiple attributes may be mentioned several times. The replacement with placeholders improves factual correctness as we replace them back with the original movie attributes later. At the end of the sentence, we append the attribute information as below:

\textbf{Original}: ``If you like La La Land, you should also see Amazing Spiderman with Emma Stone''

\textbf{With placeholder}: ``If you like {\small \texttt{[MOVIE\_TITLE\_0]}}, you should also see {\small \texttt{[MOVIE\_TITLE\_1]}} with {\small \texttt{[MOVIE\_P\_ACTOR\_0]}}; movies: La La Land (2016), The Amazing Spider-Man (2012); people: Emma Stone''

\subsection{Strategy-incorporated Model}
We prepend the strategy as a special token to the input utterance so that the model does not only generate sentences but also strategies. Similar method was used to control text generation style \cite{rashkin2019towards} as a simple and effective way to incorporate the strategies. The input to the encoder is as follows: 

\textbf{Prepend}: ``{\small \texttt{encouragement}} If you like {\small \texttt{[MOVIE\_TITLE\_0]}}, you should also see {\small \texttt{[MOVIE\_TITLE\_1]}} with {\small \texttt{[MOVIE\_P\_ACTOR\_0]}}; title: La La Land (2016), The Amazing Spider-Man (2012); people: Emma Stone''

The model first generates five candidate sentences. Then, it randomly selects a generated candidate that either contains ``encouragement'' strategy or has the greatest sentence length. In our experiment, we have tried various combinations of the top three strategies (e.g., ``encouragement'' only, ``encouragement'' and ``similarity''), and it turns out that ``encouragement'' only model gave the best result. Moreover, the sentence length selection is based on our intuition when chatting with the system. 
This aligns from our findings, ``encouragement'' is the second most frequently used strategy when humans make recommendations (\cref{strategy_def_section}), and ``recommendation'' is associated positively with successful recommendation (Table \ref{tab:movie_attribute})\footnote{We tried multi-task learning for utterance generation and strategy prediction, and the automatic metric result is comparable with the prepend method.}. 

To decide if a sentence is a recommendation or not, we train a BERT-based recommendation classifier that receives an input of recommender's current utterance and seeker's utterances from previous turn with 95.4\% accuracy and 91.2 \% F1-score.
While the index in the placeholder may become a proxy to decide whether the system needs to recommend a movie or not, it is not strictly supervised. Thus, if a generated sentence is labeled as ``recommendation'', we enforce our dialog system to recommend a new movie.

\subsection{Results}
\label{dialog_system_result}
\begin{table}
\centering
\begin{tabular}{l|cc}
\textbf{Model}& PPL$\downarrow$ & BLEU-4$\uparrow$ 
 \\ \midrule
Baseline & 9.28& 5.11 \\

Strategy & 8.93  &6.63\\
\end{tabular}
\caption{\label{tab:result} Results for automatic metrics.}
\vspace{-5mm}
\end{table} 
We compare the baseline dialog model without strategy supervision against our dialog model with strategy supervision.  We use both automatic metrics and human evaluation. 

For automatic metrics, we compute perplexity and BLEU scores \cite{papineni2002bleu}, suggesting that prepending strategies improves the model performance as shown in Table \ref{tab:result}. For human evaluation, twenty-eight participants chat with both models for 2-3 times for a more reliable judgment. We randomize which model they will chat first, in order to avoid exposure bias. After chatting, they are asked to decide which model is better in these five aspects: fluency, consistency, naturalness, persuasiveness, and engagingness. If they are unable to distinguish the dialog systems, they are allowed to choose ``can't tell'' option. 

\begin{figure}
\vspace{0mm}
\centering
{
\includegraphics[trim=2cm 3.7cm 2cm 2.4cm,clip,width=0.95\linewidth]{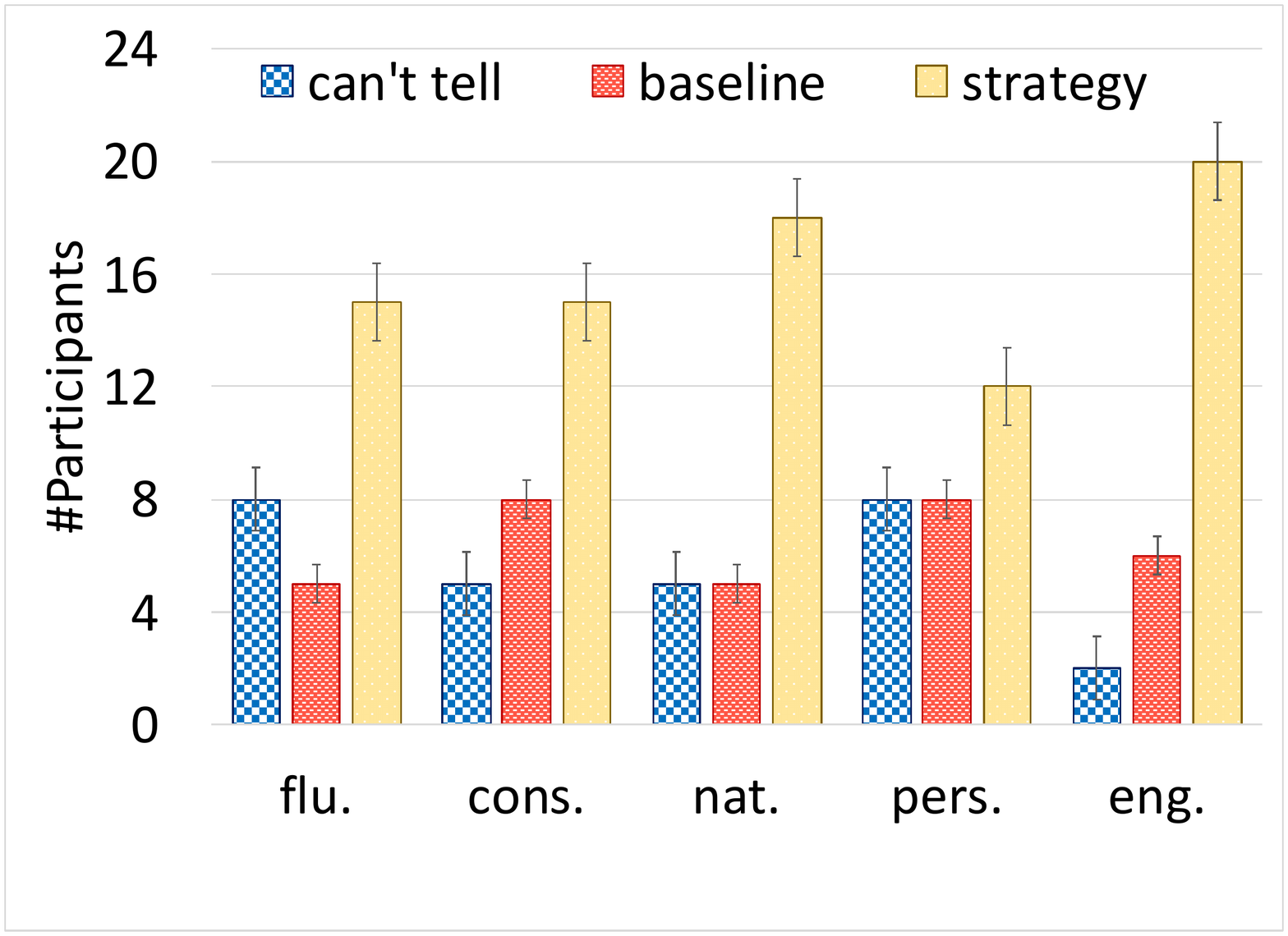}
}
\caption{\label{fig:user_study} Human evaluation result. ``Flu.'' stands for fluency, ``cons.'': consistency, ``nat.'': naturalness, ``pers.'': persuasiveness, and ``eng.'': engagingness.  
}
\vspace{-5mm}
\end{figure}

Results in Figure \ref{fig:user_study} suggest that human users prefer the model with strategy over the baseline in all aspects\footnote{We also run additional user study with five-scale ratings on these five aspects with results in Table \ref{tab:extra_user_study} in the Appendix}. It is interesting to see that although the strategy model is preferred on all metrics, people find the two model differs the most in engagingness, followed by naturalness. This supports our hypothesis that human users will find the conversations more engaging and more natural with sociable strategies incorporated in recommendation dialog systems. 

\section{Conclusion and Future Work}
In this work, we have introduced \method, a new recommendation dialog dataset collected in natural setting and annotated with sociable recommendation strategies. We analyze the connection between different strategies and the recommendation results. Our findings show that sociable strategies do have a positive impact on the acceptance of recommendation and dialog quality. This work opens up several directions for future studies in building sociable
and personalized recommendation dialog systems
as follows:

First, we will explore more ways of utilizing
the strategies, including dynamic strategy selection after decoding. Then, we plan to investigate
the strategy patterns for people with different personalities and movie preferences to make dialog
system more personalized. Finally, another interesting exploration is to extend the model with a
jointly trainable movie recommendation and movie
information modules. 

\section*{Acknowledgments}
We would like to thank members of the NLP lab at UC Davis for discussion and participation in the pilot study. We are also grateful for human evaluation participants and Mechanical Turk workers on contributions of building this dataset. 

\bibliography{anthology,emnlp2020}
\bibliographystyle{acl_natbib}
\clearpage
\begin{appendix}\label{sec:appendix}

\section{Movie Trailer Database Creation}
For each movie, we obtain metadata information from Youtube and add other movie attributes, such as plot, actors, and genre using OMBD API\footnote{\url{omdbapi.com}}.

We enrich the movies from MovieLens datasets \cite{Harper2015} with more movie trailers by searching the movie title and ``trailer''  on Youtube with a duration restriction of less than 5 minutes. We choose the trailers that are shorter than 5 minutes so that the crowd-workers do not have to spend a long time on watching them. We used the first retrieved link of the video under the duration constraint. We remove movies without a retrieved trailer from our database. Our motivation to use MovieLens and include more trailers is to link our movie database with MovieLens user review, so that it can be used for future work on building recommendation systems.

\begin{table}[h]
\centering
\begin{tabular}{lr}
\toprule
\textbf{Category} & \textbf{Coefficient}\\ \hline
\multicolumn{2}{l}{\textbf{Sociable Strategies}}\\ \hline
Personal Opinion  &  0.12*\\
Personal Experience  &0.04\\
Similarity  &  \textbf{0.24}* \\
Encouragement & 0.17*\\
Offering Help & 0.02\\
Preference Confirmation & 0.05\\
Self-Modeling  &  0.02\\ 
Credibility  & 0.07 \\
\hline
\multicolumn{2}{l}{\textbf{Preference Elicitation}} \\ \hline
Experience Inquiry & $-$0.01\\
Opinion Inquiry& 0.04\\\hline
\multicolumn{2}{l}{\textbf{Movie Attributes}} \\ \hline
Recently Released & 0.17\\
Popularity  &$-$0.10\\ \hline
\multicolumn{2}{l}{\textbf{Genre}} \\ \hline
Action & $-$0.01 \\
Romance &  $-$0.03\\
Thriller  &0.25\\
Comedy &  $-$0.06\\
Drama &  0.34\\
\bottomrule
\end{tabular}
\caption{The associations between different strategies, movie attributes, and successful recommendations. *$p <$ 0.05}
\label{tab:movie_attribute}
\end{table}


\section{Heuristics for Recommendation System}
Our heuristics for the recommendation system to handle cold-start is as follows. If the seeker never mentions a movie before and the generated text of the recommender dialog system is labeled as ``recommendation'', the most recent movie with the last mentioned genre will be recommended.

If the seeker already mentioned a movie, we will query the last mentioned movie with positive or neutral sentiment to the recommendation system for recommendation. The first recommended movie from the recommendation system output will be chosen by our dialog system. If the movie has been recommended, we will choose the next recommended movie in the recommendation output list.

To detect which movies are favored by the seekers and movie titles in the sentence, we use the modules from \citet{gunrock2020}. The sentiment classifier is a BERT-based \cite{devlin2019bert} model trained on Stanford Sentiment dataset \cite{socher2013recursive}. For the movie title detection, the model is a bidirectional LSTM-CRF with character-augmented word embedding for the input combined with retrieving similar movie title in the movie database. The movie database is from TMDB. The model was trained on speech transcripts. 

To detect movie genre in the sentence, we use regular expression matching for these genres following movie information from OMDB in our database: Action, Animation, Biography, Comedy, Crime, Drama, Documentary, Fantasy,  History, Horror, Mystery, Musical, News,  Romance, Sport, Thriller, War, and Western. To detect movie actors, actresses, and directors, we use pattern matching for capitalized first letter and find if the name exists in TMDB search for people. 

\section{Dialog Model}

The dialog model $p(d)$ of a dialog $d$ with $T$ turns is defined as follows:

\begin{equation}
    p(d) = \prod_{t=1}^{T} p_{s}(s_t|s_{<t}, r_{<t})p_{r}(r_t|s_{<t}, r_{<t})
\end{equation}

where $s$ is the seeker's utterance at turn $t$ and $r$ is the recommender's utterance at turn $t$, and $p_{s}(s_t|s_{<t}, r_{<t})$ is the probability of generating the seeker's utterance given the history. The conversation history is represented by they query/key/value features using self-attention. Interested reader can refer to \citet{ardm2019} for more details. 


\section{Training Set-up}

We adopt GPT-2 small, which is a 12-head, 12-layer, and 768-hidden size Transformer, with 117M parameters. We use pre-trained GPT-2 Byte Pair Encoding (BPE) tokenizer with the extended vocabulary of 50,310 tokens to tokenize texts. The optimizer is AdamW \cite{loshchilov2018decoupled}, and the number of warm-up steps is 100. The learning rate is set to $3\times10^{-5}$, and the dropout rate is set to 0.1. All experiments are run with an NVIDIA GeForce GTX 1080 Ti GPU.

The movie information in the input data, such as actress/actor's name, movie genre and movie plot, is delexicalized as special tokens. The real information (genre, movie title, etc.) is appended to the utterance. In addition, the strategy label is also treated as special tokens.

We leverage the ReDial \cite{li2018towards} and movie preference elicitation datasets~\cite{radlinski2019} to conduct task-related pretraining. It takes around 1.37 hour to finish one epoch in pretraining the model. 

As for the training on INSPIRED dataset, it takes around 16 minutes to finish one epoch. We train the model until it converges. The baseline model usually converges after the second epoch while the strategy-incorporated model after the third epoch. 

During the inference stage, we combined top-k based sampling and top-p based sampling (Nucleus Sampling method \cite{holtzman2019curious}). We keep the highest probability tokens whose cumulative probability mass exceeds the threshold $p$. We manually tuned the threshold of temperature, $p$ and $k$ to make both model achieve their best performance. 

The temperature is set as $0.82$ for baseline and 0.8 for the strategy-incorporated model. For both model, the threshold of $k$ is set as $400$ and the upperbound of $p$ is set as $0.9$. We manually tune the hyperparameters. 

For the strategy-incorporated model, the strategy is generated first and the utterance is then generated conditioned on the strategy. Although it is a loose constraint, the model learned categorical strategic patterns. For completeness, we also provide validation perplexity and BLEU-4 score in Table \ref{tab:val_result}

\begin{table}
\vspace{-2mm}
\centering
\begin{tabular}{l|cc|cc}
\textbf{Model}& \multicolumn{2}{c|}{\textbf{Test}} & \multicolumn{2}{c}{\textbf{Valid}} \\
&PPL & BLEU-4 & PPL & BLEU-4
 \\ \midrule
Baseline & 9.28& 5.11 & 9.21 & 5.09\\
Strategy & 8.93  &6.63 & 8.90 & 7.55\\
\end{tabular}
\caption{\label{tab:val_result} Results for automatic metrics in both validation and test data.}
\vspace{-5mm}
\end{table} 


\begin{table*}[h]

\centering
\begin{tabular}{l|ccccc}
\textbf{Model}& \textbf{Fluency} &\textbf{Consistency} & \textbf{Naturalness} & \textbf{Persuasiveness} & \textbf{Engagingness} \\ \midrule
Baseline & 3.84 & 3.80 & 3.20 & 3.24 & 3.56\\
Strategy & \textbf{4.28} & \textbf{4.16} & \textbf{3.60} & \textbf{3.52} & \textbf{3.72}\\ \hline
Human Speaker* & - & - & 4.17 & 3.97 & 4.20 \\
\end{tabular}
\caption{\label{tab:extra_user_study} Average score for human ratings on a 5-point Likert scale. Note that the human-human dialogues were collected before the user study and we did not measure fluency and consistency for human recommender.}

\end{table*} 
\section{Additional User Study}
In addition to the comparison study done by human users mentioned in \cref{dialog_system_result}, we conduct another user study which asks each participant to rate from 1 (worst) to 5 (best) for the same five aspects: fluency, consistency, naturalness, persuasiveness, and engagingness. For each model, 25 participants chat interactively with it (in total: 50 users). Unlike the user study in \cref{dialog_system_result} where 1 user interacts with both models, this time a user interacts with one model since users do not need to compare. These participants are different from the ones reported in the comparison user study (\cref{dialog_system_result}). From Table \ref{tab:extra_user_study}, we can see that the strategy model has higher ratings than the baseline model in all aspects.

\section{Example Human-Human Dialogs in \textsc{Inspired}}
We include 2 annotated examples of human-human dialogs in Table \ref{tab:example_human_human_dialogs} and \ref{tab:example2_human_human_dialogs}.

\section{Example Human-System Dialogs}
We include example dialog of human seeker and the baseline model in Table \ref{tab:baseline_bot} and an example strategy-incorporated dialog model in Table \ref{tab:strategy_bot} from user study. In the user study of evaluating the dialog system, we do not set a minimum turn for the human user. 

\section{Dialog Collection and Annotation Interfaces}
\label{interfaces}
\begin{figure*}[h]
\vspace{0mm}
\centering
{
\includegraphics[width=0.7\linewidth]{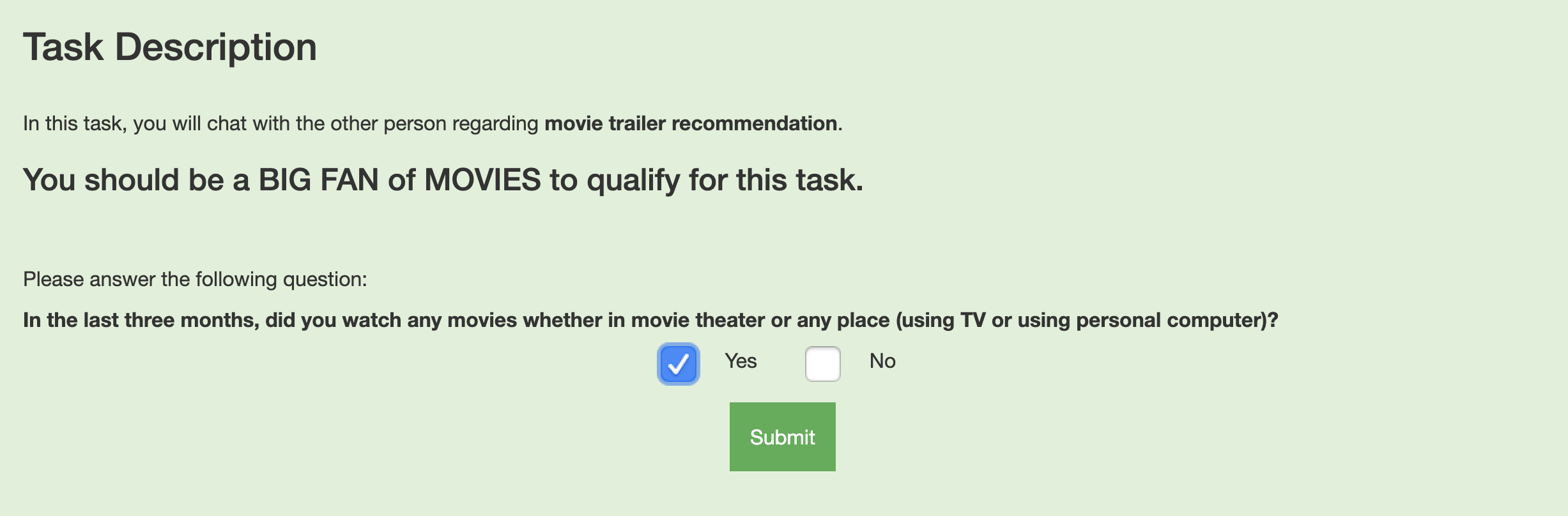}
}

\caption{\label{fig:bigfan} Onboarding page of both seekers and recommenders}
\vspace{0mm}
\end{figure*}

\begin{figure*} [h]
\vspace{0mm}
\centering
{
\includegraphics[width=0.7\linewidth]{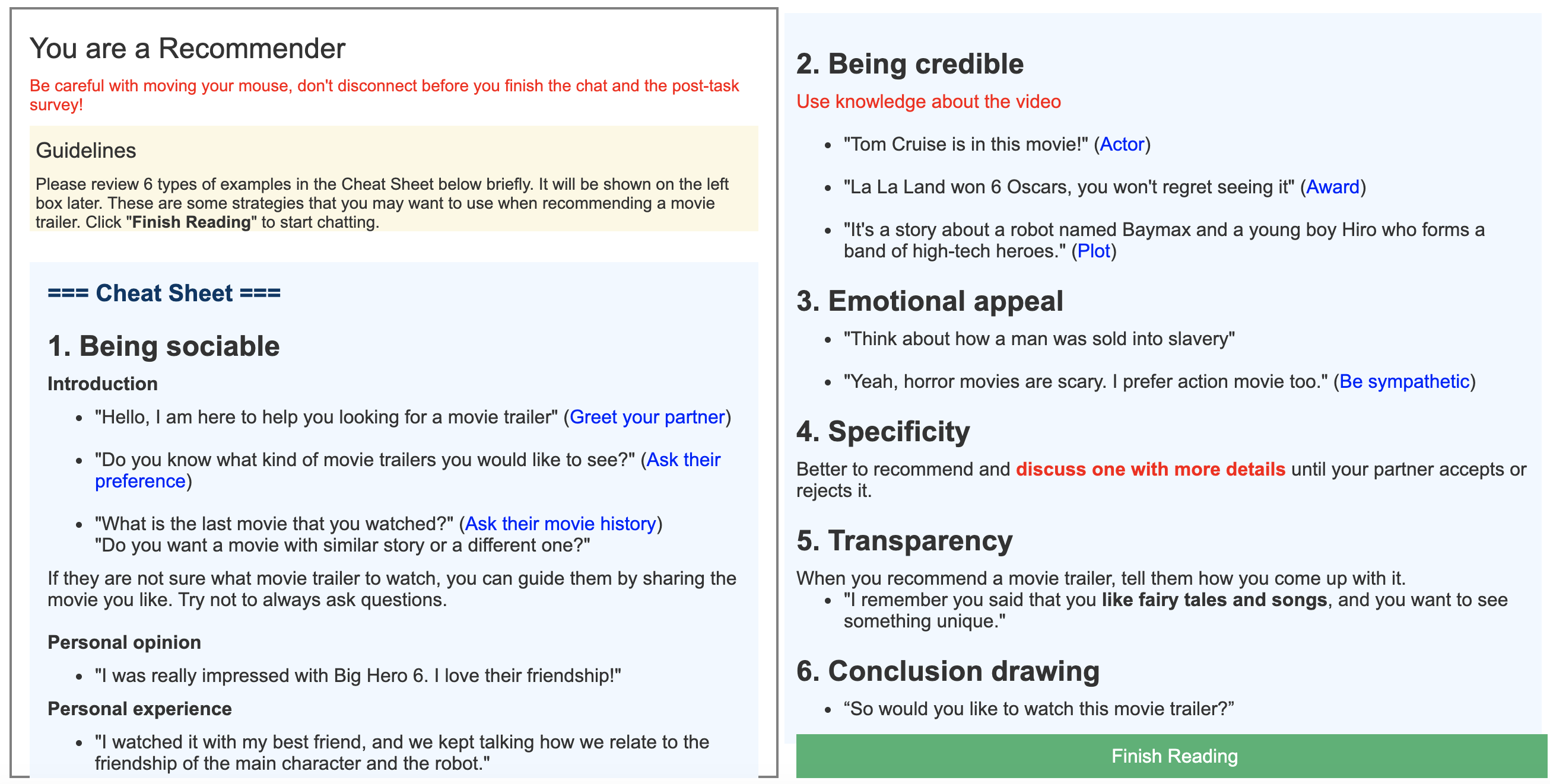}
}
\caption{\label{fig:cheat_sheet} Strategy tips for recommenders before they start chatting}
\vspace{0mm}
\end{figure*}

\begin{figure*}
\vspace{0mm}
\centering
{
\includegraphics[width=0.7\linewidth]{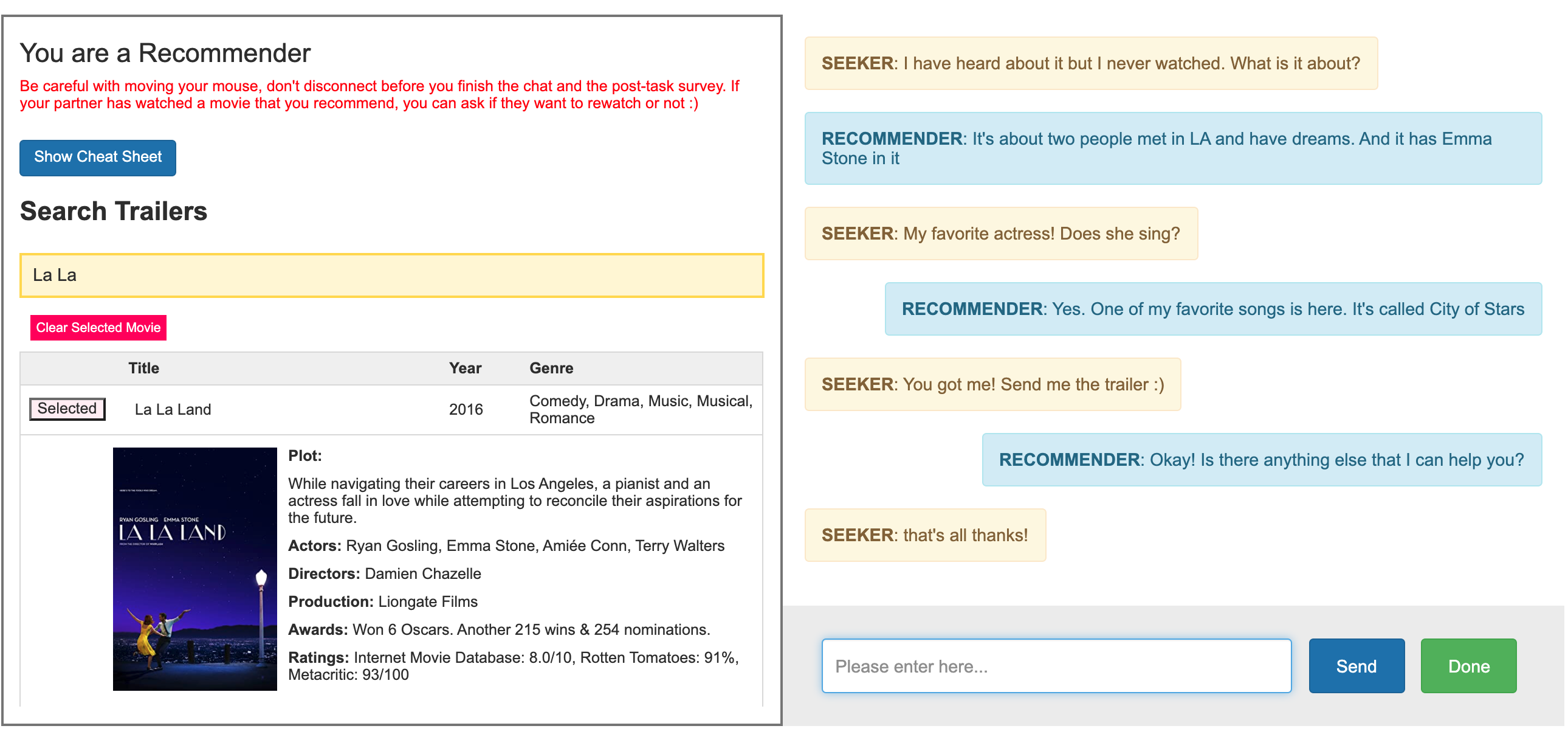}
}
\caption{\label{fig:rec_chat} Recommender's chat interface during recommendation phase. On the left side is the search bar.}
\vspace{0mm}
\end{figure*}

\begin{figure*}[h]
\vspace{0mm}
\centering
{
\includegraphics[width=0.7\linewidth]{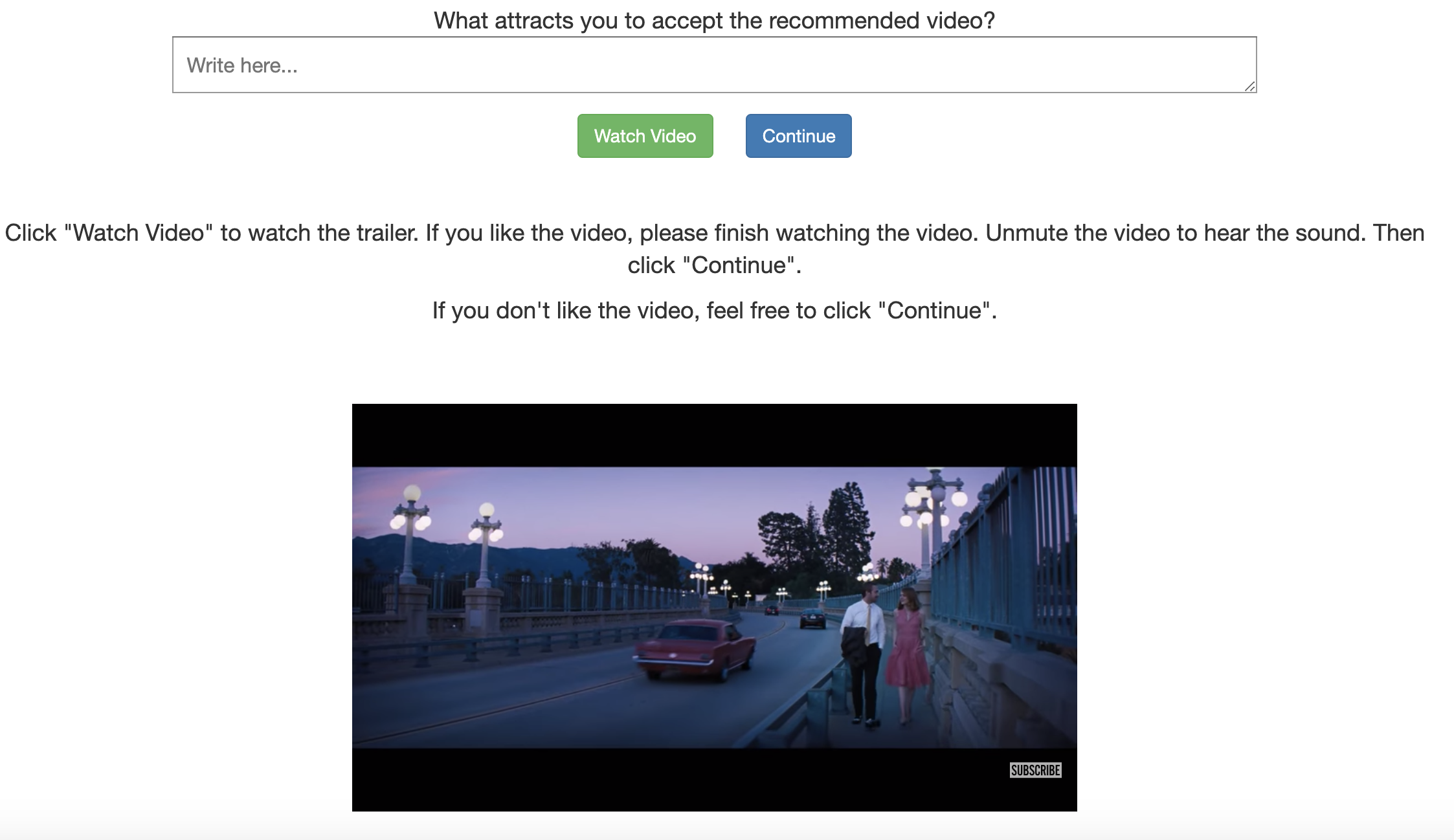}
}

\caption{\label{fig:seeker_watch_video}  A page for seeker to watch the recommended video}
\vspace{0mm}
\end{figure*}

\begin{figure*}[h]
\vspace{0mm}
\centering
{
\includegraphics[trim=1.3cm 0cm 1cm 0cm,clip,width=\linewidth]{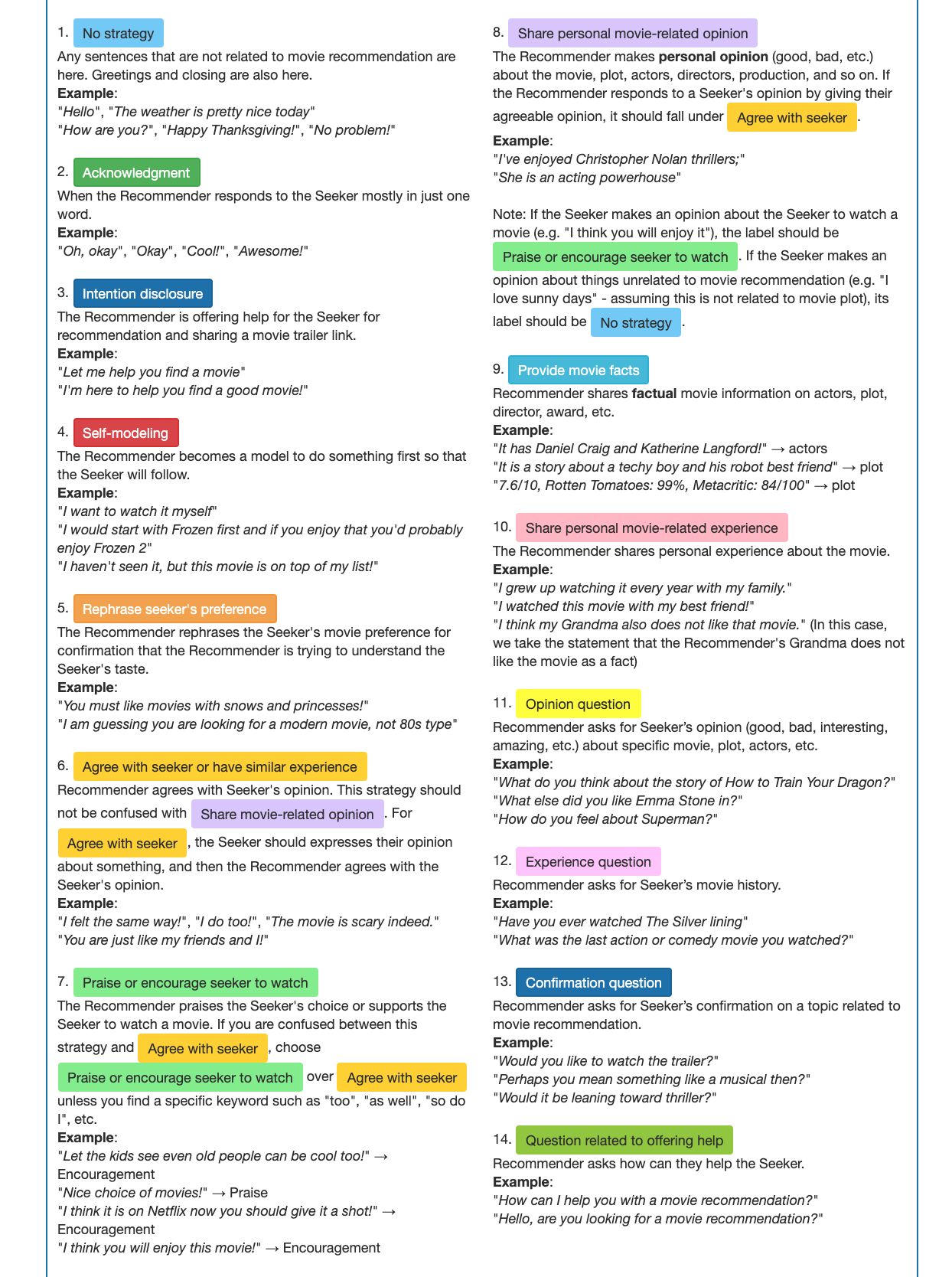}
}

\caption{\label{fig:strategy_definition} A screenshot of the website interface describing strategy definition for annotation task for the crowd-workers}
\vspace{0mm}
\end{figure*}

\begin{figure*}[h]
\vspace{0mm}
\centering
{
\includegraphics[width=\linewidth]{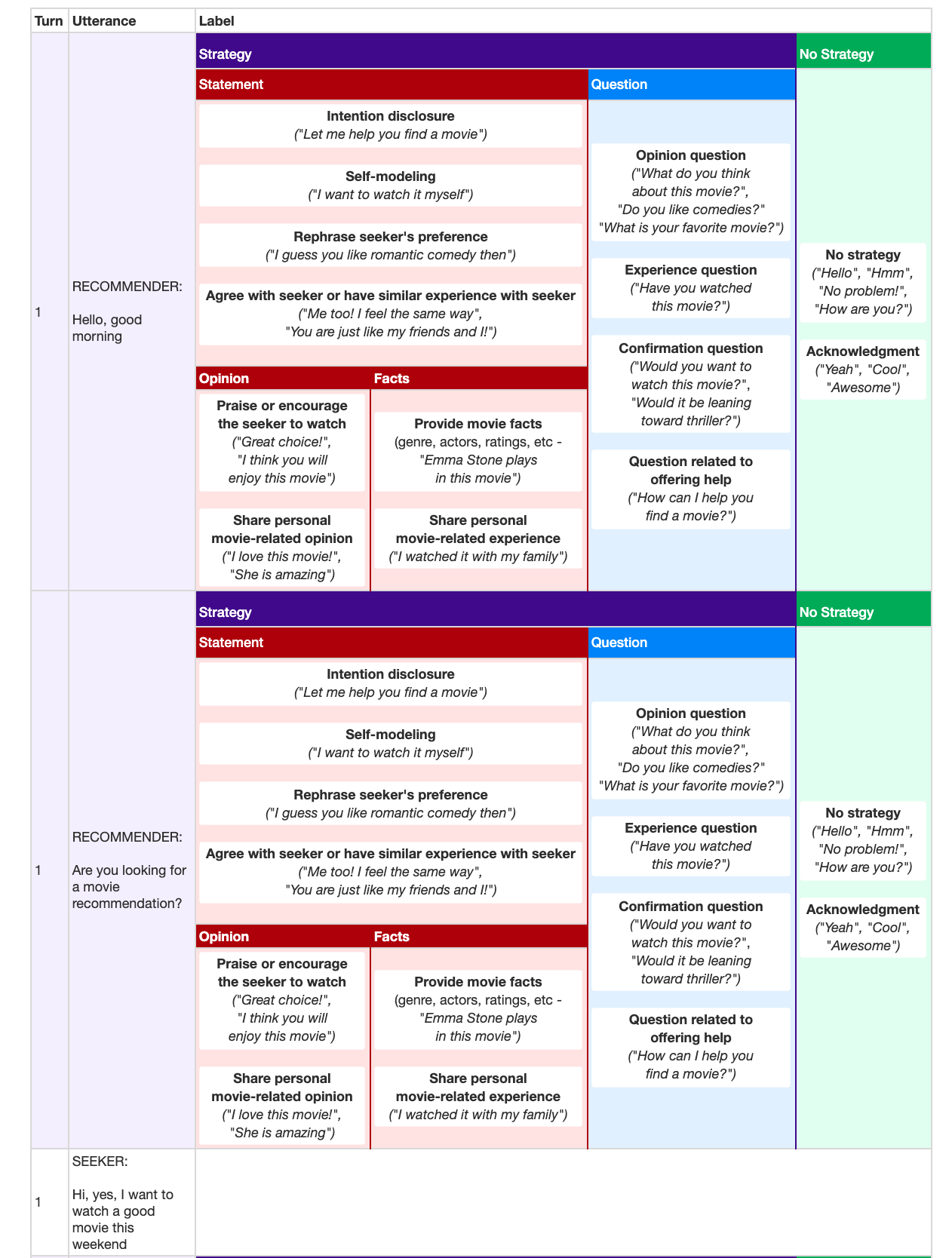}
}
\caption{\label{fig:interface} Interface for crowd-workers to annotate the dialogs}
\vspace{0mm}
\end{figure*}

Figure \ref{fig:bigfan}, 
\ref{fig:cheat_sheet},
\ref{fig:rec_chat},
\ref{fig:seeker_watch_video} show dialog collection interface. 
Figure \ref{fig:strategy_definition} and \ref{fig:interface} are dialog annotation interfaces for the crowd-workers. 

\begin{table*}[h]
\centering
\small
\begin{tabular}{lll}
\toprule
\multicolumn{2}{l}{\textbf{Conversation 1}} & \textbf{Label}\\ \midrule
\textcolor{red}{REC}: & Hi There!  &  \texttt{No Strategy}\\
& What types of movies  do you like to watch?& \texttt{Offering Help} \\
\textcolor{blue}{SEEK}: & Hello! I'm more of an action movie or a good romance and mystery movie.\\
\textcolor{red}{REC}: & I just saw the trailer for Knives Out when I went to see Joker and it looked like & \texttt{Self-Modeling}\\
& a good mix of action and mystery! \\
\textcolor{blue}{SEEK}: & I seen that one too as I seen Joker about a month ago. &\\
& I thought about asking my fiance about going and seeing it.&\\
\textcolor{red}{REC}: & It looks like a good movie for people who like many different movies. & \texttt{Personal Opinion}\\
& It also has a great cast! & \texttt{Personal Opinion}\\
& I was surprised to see Chris Evans in the trailer! & \texttt{Personal Opinion}\\
\textcolor{blue}{SEEK}: & Maybe with Chris Evans in it it'll be easier to convince my fiance to see it.\\
& Do you know who else is in the cast?\\
\textcolor{red}{REC}: & Daniel Craig and Jamie Lee Curtis are also in the cast. &\texttt{Credibility}\\
& Daniel Craig does a lot of 007 so definitely a good hearthrob role  & \texttt{Encouragement}\\
& to convince the misses lol!\\
\textcolor{blue}{SEEK}:& I am the misses lol. But he loves the bond movies so that should be a good \\
&  incentive for him to go see it. Do you have any other recommendations? \\
\textcolor{red}{REC}: & The new Star Wars comes out in less than a month, if you are into the franchise.\\
\textcolor{blue}{SEEK}:&  He is, I think he told me we're getting it when it comes out to add to our movie \\
& collection.\\
\textcolor{red}{REC}: & Well that is another great action movie. & \texttt{Encouragement}\\
& I also recommend the John Wick series & \texttt{Encouragement}\\
\textcolor{blue}{SEEK}:& I haven't seen any of that series. Could you tell me what the general plot is$>$\\
\textcolor{red}{REC}: & John Wick is a former member of a gang, he was basically an assassin. & \texttt{Credibility} \\
& He falls in love and quits the game, but then his wife dies, and someone comes in & \texttt{Credibility} \\
&  and kills his dog. \\
& He then goes on a revenge rampage against the people who broke into his house. & \texttt{Credibility} \\
& I have yet to watch the 3rd one but the action scenes were really cool! & \texttt{Personal Opinion}\\
\textcolor{blue}{SEEK}:&  Oh I'd definitely would cry at the dogs death. \\
\textcolor{red}{REC}: & It is really sad! & \texttt{Similarity}\\
& the dog was a last gift from his dying wife which makes it so much worse & \texttt{Personal Opinion}\\
\textcolor{blue}{SEEK}:& I couldn't even finish I am legend because of the dog dying.\\
& Anything with animal death makes me ball like a baby.\\
\textcolor{red}{REC}: & Marley \& Me had me crying for a good half hour so I completely understand that! & \texttt{Similarity}\\
\textcolor{blue}{SEEK}:&I avoided that movie because someone told me he passed away. \\
& My fiance took me to see jurrasic world as our first date and I cried at the \\
& dinosuars dying.\\
\textcolor{red}{REC}: &I would definitely avoid that movie if animal deaths make you said. &\texttt{Similarity}\\
& Oh that is so cute though! &\texttt{No Strategy}\\
\textcolor{blue}{SEEK}:& Yeah, he had to calm me down for about an hour and bought me ice cream to  \\
& apologize for it.\\
\textcolor{red}{REC}: & Aww that is so sweet.&\texttt{No Strategy}\\
& Given that you dont want to see animals die, and you are looking for an  & \texttt{Pref. Confirmation}\\
&  Action$/$Mystery, I think Knives Out would be a good movie choice. \\
& Do you agree?& \texttt{Pref. Confirmation}\\
\textcolor{blue}{SEEK}:& I do agree with that. When it comes out i'll bring it up for date night. Thank you!!\\
\bottomrule
\end{tabular}
\caption{Example 1 for labeled human-human dialog. REC refers Recommender and SEEK to Seeker.}
\label{tab:example_human_human_dialogs}
\end{table*}

\begin{table*}[h]
\centering
\small
\begin{tabular}{lll}
\toprule
\multicolumn{2}{l}{\textbf{Conversation 2}} & 
\textbf{Label}\\ \midrule
\textcolor{red}{REC}: & Hi there. & \texttt{No Strategy}\\
& How are you? & \texttt{No Strategy}\\
\textcolor{blue}{SEEK}:& I'm doing well. I was wondering if you had any movie recommendations. \\
\textcolor{red}{REC}: & Let me see. & \texttt{Offering Help}\\
& Well, what kind of movies do you usually find yourself watching? & \texttt{Opinion Inquiry}\\
& What do you like? & \texttt{Opinion Inquiry}\\
\textcolor{blue}{SEEK}:& I tend to like action movies that also involve comedy.\\
\textcolor{red}{REC}: & Oh, me too. & \texttt{Similarity} \\
& So, what did you see last that you liked? & \texttt{Experience Inquiry}\\
\textcolor{blue}{SEEK}:& I really liked Avengers: Endgame because I liked the pacing and\\
& the effects were great.\\
\textcolor{red}{REC}: & Okay, I think I kind of understand your taste a little. & \texttt{Pref. Confirmation}\\
& It's sort of similar to mine, although I haven't seen an action  & \texttt{Similarity}\\
& hero movie in a while.\\
& Did you like Guardians of the Galaxy? & \texttt{Opinion Inquiry}\\
\textcolor{blue}{SEEK}:& No, I haven't, but I heard good things about it.\\
\textcolor{red}{REC}: & Oh, I was assuming that you'd seen it. & \texttt{Offering Help}\\
& I saw that with my son and I was surprised at how much I liked it. & \texttt{Personal Experience}\\
\textcolor{blue}{SEEK}:& Were you able to see the sequel for it?\\
\textcolor{red}{REC}: &No, but I would like to. & \texttt{Self-Modeling}\\
& I would really recommend that. & \texttt{Encouragement}\\
& I don't know if you know who Chris Pratt is, but he's a keeper. & \texttt{Personal Opinion}\\
& He was on Parks \& Rec, the comedy show. & \texttt{Credibility}\\
\textcolor{blue}{SEEK}:& I love Chris Pratt! I accept your recommendation. Thank you so much.\\
\textcolor{red}{REC}: & :) oh, I'm so happy. & \texttt{Acknowledgment}\\
& Cool. & \texttt{Acknowledgment}\\
& I hope you like it. & \texttt{Encouragement}\\
\textcolor{blue}{SEEK}:&Do you have any other recommendations for action movies? \\
\textcolor{red}{REC}: & Hmmm...I don't know if I can think of any  others that are funny like that. & \texttt{No Strategy} \\
& I know there are some though. & \texttt{No Strategy}\\
\textcolor{blue}{SEEK}:& Oh, I just remembered. I really liked Deadpool because it did a great job of \\
& mashing comedy and action together.\\
\textcolor{red}{REC}: & My 20 year old son liked that. & \texttt{Personal Experience}\\
& I have only seen a couple parts.  & \texttt{Personal Experience}\\
& You've no doubt seen the MIB series, but those were kind of good as well. & \texttt{Personal Opinion}\\
\textcolor{blue}{SEEK}:& Can you tell me what you like about MIB?\\
& I like Will Smith, but I haven't see any of the movies.\\
\textcolor{red}{REC}: & Oh, yeah. & \texttt{Acknowledgment}\\
& They're telling us to quit the hit, but real quick. & \texttt{No Strategy}\\
& He's funny and creative in it and he's great when paired with Tommy Lee Jones. & \texttt{Personal Opinion}\\
& It's just creative and makes you laugh unexpectedly. & \texttt{Personal Opinion}\\
& Anyway, that's all I got. & \texttt{No Strategy}\\
& It was nice talking to you. & \texttt{No Strategy}\\
\textcolor{blue}{SEEK}:&It was great talking to you.  I'll be sure to check out your recommendations.\\
\bottomrule
\end{tabular}
\caption{Example 2 for labeled human-human dialog. REC refers Recommender and SEEK to Seeker.}
\label{tab:example2_human_human_dialogs}
\end{table*}

\begin{table*}[h]
\centering
\small
\begin{tabular}{ll}
\toprule
\multicolumn{2}{l}{\textbf{Human Conversation with Baseline Dialog Model}} \\ \midrule
\textcolor{red}{REC\_SYS}: & Hello there!\\
\textcolor{blue}{SEEK}:& hello, how are you\\
\textcolor{red}{REC\_SYS}: &I'm doing good!\\
\textcolor{blue}{SEEK}:& that's nice to hear\\
\textcolor{red}{REC\_SYS}: & I'm here to help you look for a movie trailer.\\
\textcolor{blue}{SEEK}:& really, do you have a recommendation?\\
\textcolor{red}{REC\_SYS}: & Do you know what kind of movie trailers you'd like to see?\\
\textcolor{blue}{SEEK}:& I like super hero movies\\
\textcolor{red}{REC\_SYS}: & Do you know what kind of movies you'd like to see?\\
\textcolor{blue}{SEEK}:& i like action movies\\
\textcolor{red}{REC\_SYS}: & I do too.\\
\textcolor{blue}{SEEK}:& that's pretty awesome, I am glad we have something in common\\
\textcolor{red}{REC\_SYS}: &Have you seen the movie Avengers: Endgame?\\
\textcolor{blue}{SEEK}:& yes\\
\textcolor{red}{REC\_SYS}: &What about Avengers: Endgame?\\
\textcolor{blue}{SEEK}:& yes I have seen it\\
\textcolor{red}{REC\_SYS}: & Have you seen the movie Assassin's Creed?\\
\textcolor{blue}{SEEK}:& no i haven't\\
\textcolor{red}{REC\_SYS}: &Did you enjoy it?\\
\textcolor{blue}{SEEK}:& i don't know because I haven't seen it\\
\textcolor{red}{REC\_SYS}: &If you liked it, you'll love the movie Assassin's Creed.\\
\textcolor{blue}{SEEK}:& hm, that's great.\\
\textcolor{red}{REC\_SYS}: &Great!\\
\bottomrule
\end{tabular}
\caption{Example dialog of human-system. REC\_SYS refers baseline recommendation dialog system and SEEK to human Seeker.}
\label{tab:baseline_bot}
\end{table*}

\begin{table*}[h]
\centering
\small
\begin{tabular}{ll}
\toprule
\multicolumn{2}{l}{\textbf{Human Conversation with Strategy-Incorporated Dialog Model}} \\ \midrule
\textcolor{red}{REC\_SYS}: & Hey there, how are you today?\\
\textcolor{blue}{SEEK}:& Hello. I'm good. How are you? \\
\textcolor{red}{REC\_SYS}: & Great, thank you. I am great. What kind of movies do you like to watch?\\
\textcolor{blue}{SEEK}:&I like to watch romantic comedies. What should I watch today?\\
\textcolor{red}{REC\_SYS}: &Awesome. I love comedy movies as well. What is your favorite comedy movie? \\
\textcolor{blue}{SEEK}:& My favourite comedy movie is called 3 idiots\\
\textcolor{red}{REC\_SYS}:&Yes I love that movie! How about Like a Boss?\\ 
\textcolor{blue}{SEEK}:&I've never watched that before\\
\textcolor{red}{REC\_SYS}: & Awesome! It is a comedy movie. \\
& Do you prefer movies that have a little bit of humor or are more of an action/adventure type of movie?\\
\textcolor{blue}{SEEK}:& A bit of humor is goodd\\
\textcolor{red}{REC\_SYS}: & Onward is a good one for sure. It is a fun comedy.\\
\textcolor{blue}{SEEK}:& What is that one about?\\
\textcolor{red}{REC\_SYS}:& "In a suburban fantasy world, two teenage elf brothers embark on an extraordinary quest to discover \\
& if there is still a little magic left out there."\\
\textcolor{blue}{SEEK}:& Oh that's cool. I might watch that one then\\
\textcolor{red}{REC\_SYS}: & Onward is a great movie to see if you are into comedy movies. You should definitely watch it. \\
\bottomrule
\end{tabular}
\caption{Example dialog of human-system. REC\_SYS refers strategy-incorporated recommendation dialog system and SEEK to human Seeker.}
\label{tab:strategy_bot}
\end{table*}
\clearpage

\begin{figure*}[h]
\vspace{0mm}
\centering
{
\includegraphics[width=0.7\linewidth]{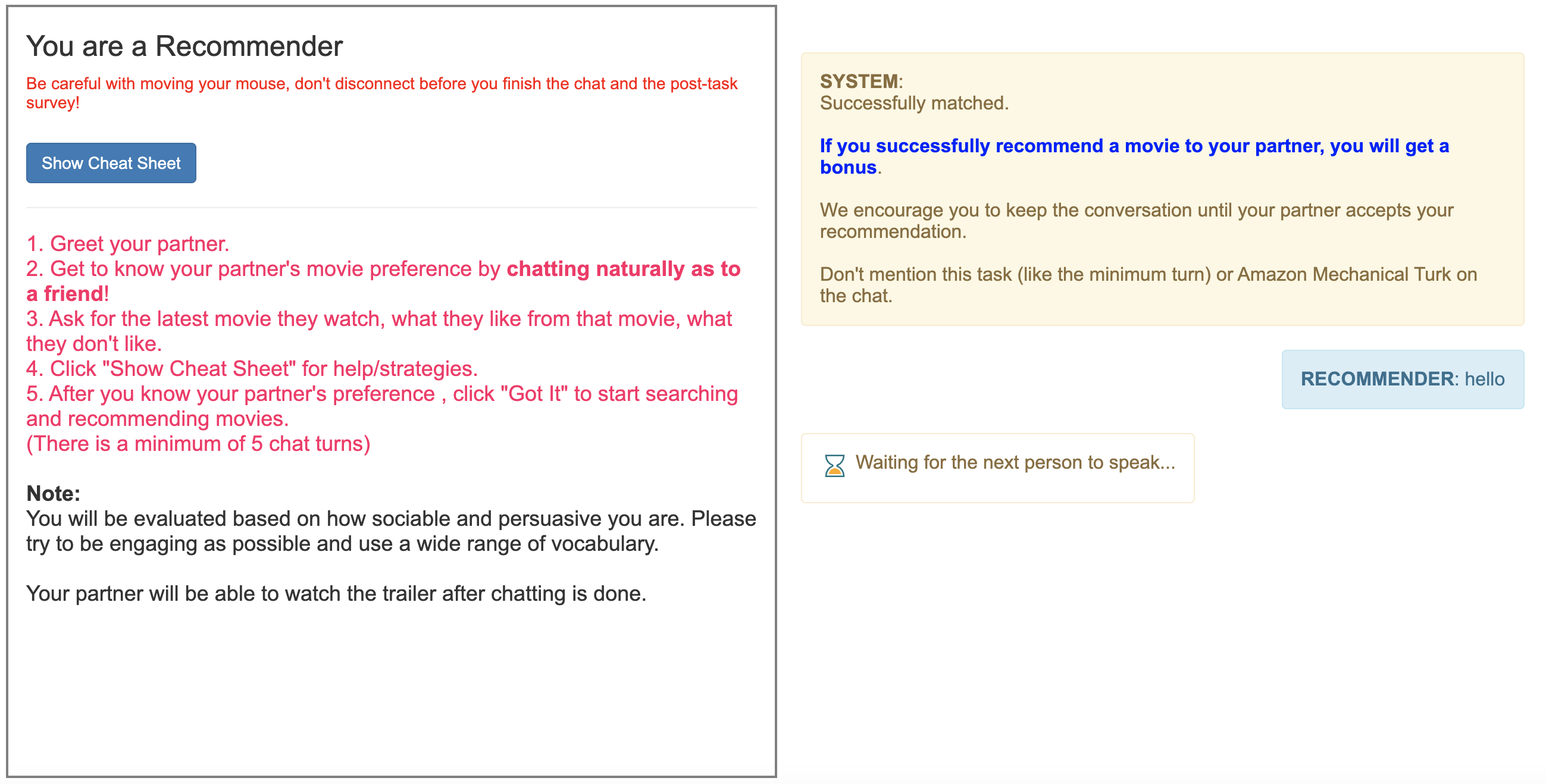}
}
\caption{\label{fig:inquiry_chat} Recommender's chat interface during preference elicitation phase}
\vspace{0mm}
\end{figure*}

\begin{figure*}
\vspace{0mm}
\centering
{
\includegraphics[width=0.9\linewidth]{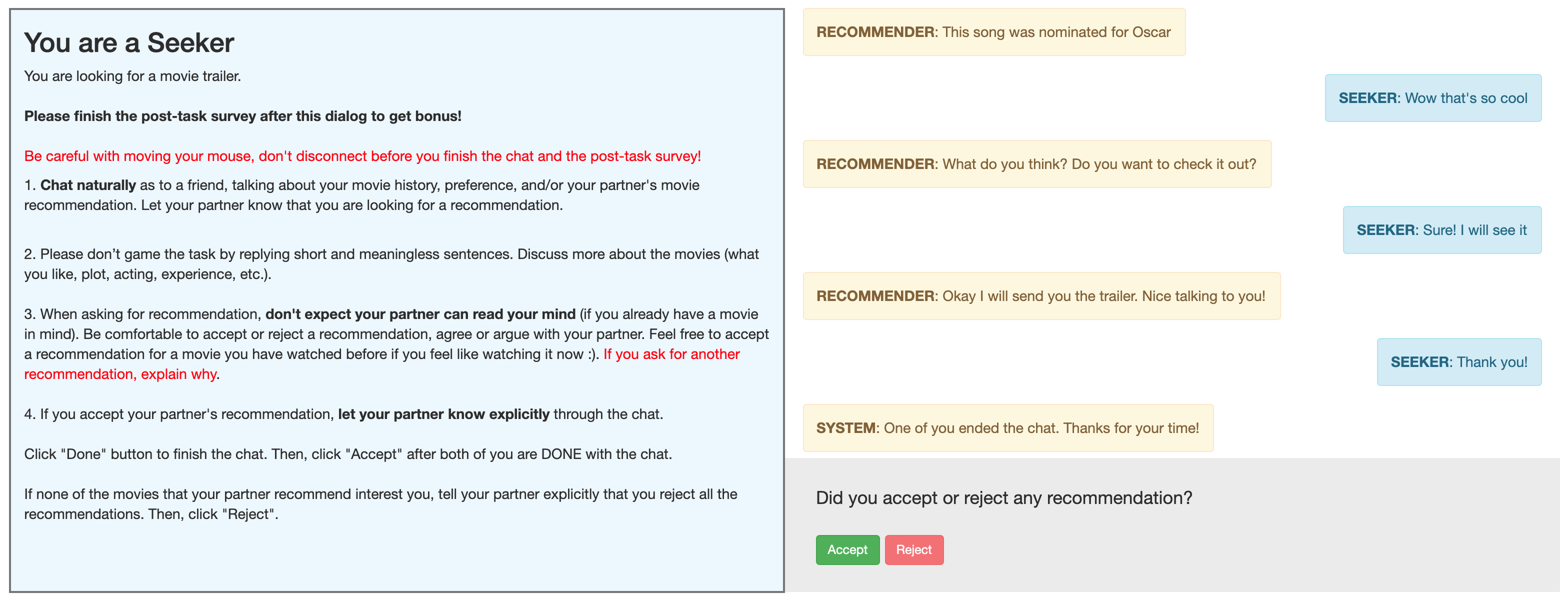}
}
\caption{\label{fig:seeker_chat} Seeker's chat interface after they finish the chat, and the Recommender sends over the movie. Seeker can choose to accept or reject.}
\vspace{0mm}
\end{figure*}

\end{appendix}
\end{document}